\documentclass[a4paper,twoside]{article}

\usepackage{epsfig}
\usepackage{subcaption}
\usepackage{calc}
\usepackage{amssymb}
\usepackage{amstext}
\usepackage{amsmath}
\usepackage{amsthm}
\usepackage{multicol}
\usepackage{pslatex}

\usepackage{multirow}
\usepackage{natbib}

\usepackage{SCITEPRESS}     

\newcommand\blfootnote[1]{%
  \begingroup
  \renewcommand\thefootnote{}\footnote{#1}%
  \addtocounter{footnote}{-1}%
  \endgroup
}

\newcommand{\ssymbol}[1]{^{\@fnsymbol{#1}}}

\begin{document}

\title{Video Camera Identification from Sensor Pattern Noise with a Constrained ConvNet}
\author{\authorname{Derrick Timmerman\textsuperscript{*}
\sup{1}\orcidAuthor{0000-0002-9797-8261}, Guru Swaroop Bennabhaktula\textsuperscript{*}\sup{1}\orcidAuthor{0000-0002-8434-9271}, Enrique Alegre\sup{2}\orcidAuthor{0000-0003-2081-774X}, and George Azzopardi\sup{1}\orcidAuthor{0000-0001-6552-2596}}
\affiliation{\sup{1}Bernoulli Institute for Mathematics, Computer Science and Artificial Intelligence, \\University of Groningen, The Netherlands}
\affiliation{\sup{2} Group for Vision and Intelligent Systems, Universidad de Le\'on, Spain}
\email{d.k.timmerman@rug.nl, g.s.bennabhaktula@rug.nl, g.azzopardi@rug.nl, enrique.alegre@unileon.es}
}


\abstract{
The identification of source cameras from videos, though it is a highly relevant forensic analysis topic, has been studied much less than its counterpart that uses images. In this work we propose a method to identify the source camera of a video based on camera specific noise patterns that we extract from video frames. For the extraction of noise pattern features, we propose an extended version of a constrained convolutional layer capable of processing color inputs. Our system is designed to classify individual video frames which are in turn combined by a majority vote to identify the source camera. We evaluated this approach on the benchmark VISION data set consisting of $1539$ videos from $28$ different cameras. To the best of our knowledge, this is the first work that addresses the challenge of video camera identification on a device level. The experiments show that our approach is very promising, achieving up to $93.1\%$ accuracy while being robust to the WhatsApp and YouTube compression techniques. This work is part of the EU-funded project 4NSEEK focused on forensics against child sexual abuse.
}

\onecolumn \maketitle \normalsize \setcounter{footnote}{0} \vfill

\section{\uppercase{Introduction}}
\label{sec:introduction}
\blfootnote{\textsuperscript{*}Derrick Timmerman and Guru Swaroop Bennabhaktula are both first authors.}

\noindent
Source camera identification of digital media plays an important role in counteracting problems that come along with the simplified way of sharing digital content. Proposed solutions aim to reverse-engineer the acquisition process of digital content to trace the origin, either on a \textit{model} or \textit{device} level. Whereas the former aims to identify the brand and model of a camera, the latter aims at identifying a specific instance of a particular model. Detecting the source camera that has been used to capture an image or record a video can be crucial to point out the actual owner of the content, but could also serve as additional evidence in court.

Proposed techniques typically aim to identify the source camera by extracting \textit{noise patterns} from the digital content. Noise patterns can be thought of as an invisible trace, intrinsically generated by a particular device. These traces are the result of imperfections during the manufacturing process and are considered unique for an individual device \citep{lukavs2006detecting}. By its unique nature and its presence on every acquired content, the noise pattern functions as an instrument to identify the camera model. 

Within the field of source camera identification a distinction is made between image and video camera identification. Though a digital camera can capture both images and videos, in a recent study it is experimentally shown that a system designed to identify the camera model of a given image cannot be directly applied to the problem of video camera model identification \citep{hosler2019video}. Therefore, separate techniques are required to address both problems.

Though great effort is put into identifying the source camera of an image, significantly less research has been conducted so far on a similar task using digital videos \citep{milani2012overview}. Moreover, to the best of our knowledge, only a single study addresses the problem of video camera model identification by utilizing deep learning techniques \citep{hosler2019video}. Given the potential of such techniques in combination with the prominent role of digital videos in shared digital media, the main goal of this work is to further explore the possibilities of identifying the source camera at device level of a given video by investigating a deep learning pipeline.

In this paper we present a methodology to identify the source camera device of a video.
We train a deep learning system based on the constrained convolutional neural network architecture proposed by \citet{bayar2018constrained} for the extraction of noise patterns, to classify individual video frames. Subsequently, we identify the video camera device by applying the simple majority vote after aggregating frame classifications per video.

With respect to current state-of-the-art approaches, we advance with the following contributions: i) to the best of our knowledge, we are the first to address video camera identification on a device level by including multiple instances of the same brand and model; ii) we evaluate the robustness of the proposed method with respect to common video compression techniques for videos shared on the social media platforms, such as WhatsApp and YouTube; iii) we propose a multi-channel constrained convolutional layer, and conduct experiments to show its effectiveness in extracting better camera features by suppressing the scene content. 

The rest of the paper is organized as follows. We start by presenting an overview of model-based techniques in source camera identification, followed by current state-of-the-art approaches in Section~\ref{sec:related_work}. In Section~\ref{sec:methodology} we describe the methodology for the extraction of noise pattern features for the classification of frames and videos. Experimental results along with the data set description are provided in Section~\ref{sec:exp-results}. We provide a discussion of certain aspects of the proposed work in Section~\ref{sec:discussion} and finally, we draw conclusions in Section~\ref{sec:conclusion}.

\section{\uppercase{Related Work}}
\label{sec:related_work}
\begin{figure}[t]
  \centering
    {\epsfig{file = 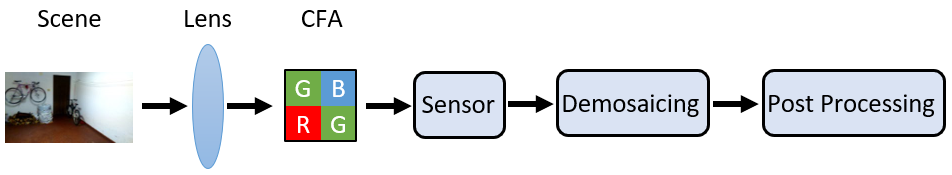, width = 7.5cm}}
  \caption{Acquisition pipeline of an image. Adapted from \cite{chen2015camera}.}
  \label{fig:imgacq}
\end{figure}

\noindent 
In the past decades, several approaches have been proposed to address the problem of image camera model identification \citep{bayram2005source, li2010source, bondi2016first, icpram20}. Those methodologies aim to extract noise pattern features from the input image or video that characterise the respective camera model. These noise patterns or traces are the result of imperfections during the manufacturing process and are thought to be unique for every camera model \citep{lukavs2006detecting}. More specifically, during the acquisition process at shooting time, camera models perform series of sophisticated operations applied to the raw content before it is saved in memory, as shown in Fig.~\ref{fig:imgacq}. During these operations, characteristic traces are introduced to the acquired content, resulting in a unique noise pattern embedded in the final output image or video. This noise pattern is considered to be deterministic and irreversible for a single camera sensor and is added to every image or video the camera acquires \citep{caldell2010multimedia}.

\subsection{Model-based Techniques}
Based on the hypothesis of unique noise patterns, many image camera model identification algorithms have been proposed aiming at capturing these characteristic traces which can be divided into two main categories: hardware and software based techniques. Hardware techniques consider the physical components of a camera such as the camera's CCD (Charge Coupled Device) sensor \citep{geradts2001methods} or the lens \citep{dirik2008digital}. Software techniques capture traces left behind by internal components of the acquisition pipeline of the camera, such as the sensor pattern noise (SPN) \citep{lukas2006digital} or demosaicing strategies \citep{milani2014demosaicing}.

\subsection{Data-driven Technologies}
Although model-based techniques have shown to achieve good results, they all rely on manually defined procedures to extract (parts of) the characteristic noise patterns. Better results are achieved by applying deep learning techniques, also known as data-driven methodologies. There are a few reasons why these methods work so well. First, these techniques are easily scalable since they learn directly from data. Therefore, adding new camera models does not require manual effort and is a straightforward process. Second, these techniques often perform better when trained with large amounts of data, allowing us to take advantage of the abundance of digital images and videos publicly available on the internet. 

Given their ability to learn salient features directly from data, convolutional neural networks (ConvNets) are frequently incorporated to address the problem of image camera model identification. To further improve the feature learning process of ConvNets, tools from steganalysis have been adapted that suppress the high level scene content of an image \citep{qiu2014universal}. In their existing form, convolutional layers tend to extract features that capture the scene content of an image as opposed to the desired characteristic camera detection features, i.e. the noise patterns. This behavior was first observed by \citet{chen2015median} in their study to detect traces of median filtering. Since the ConvNet was not able to learn median filtering detection features by feeding images directly to the input layer, they extracted the median filter residual (i.e. a high dimensional feature set) from the image and provided it to the ConvNet's input layer, resulting in an improvement in classification accuracy.

Following the observations of \cite{chen2015median}, two options for the ConvNet have emerged that suppress the scene content of an image: using a predetermined high-pass filter (HPF) within the input layer \citep{pibre2016deep} or the adaptive constrained convolutional layer \citep{bayar2016deep}. Whereas the former requires human intervention to set the predetermined filter, the latter is able to jointly suppress the scene content and to adaptively learn relationships between neighbouring pixels. Initially designed for image manipulation detection, the constrained convolutional layer shows to achieve state-of-the-art results in other digital forensic problems as well, including image camera model identification \citep{bayar2017design}.

Although deep learning techniques are commonly used to identify the source camera of an image, it was not until very recently that these techniques were applied to video camera identification. Therefore, the work of \citet{hosler2019video} is one of the few ones closely related to the ideas we propose. \citeauthor{hosler2019video} adopted the constrained ConvNet architecture proposed by \cite{bayar2018constrained}, although they removed the constrained convolutional layer due to its incompatibility with color inputs. To identify the camera model of a video, in their work they train the ConvNet to produce classification scores for patches extracted from video frames. Subsequently, individual patch scores are combined to produce video-level classifications, as single-patch classifications showed to be insufficiently reliable.

The ideas we propose in this work differ from \cite{hosler2019video} in the following ways. Instead of removing the constrained convolutional layer, we propose an extended version of it by making it suitable for color inputs. Furthermore, instead of using purely different camera models, we include $28$ camera devices among which $13$ are of the same brand and model, allowing us to investigate the problem in a device-based manner. Lastly, we provide the network with the entire video frame to extract noise pattern features, whereas \citeauthor{hosler2019video} use smaller patches extracted from frames. 
\section{\uppercase{Methodology}}
\label{sec:methodology}
\begin{figure}[t]
  \centering
    {\epsfig{file = 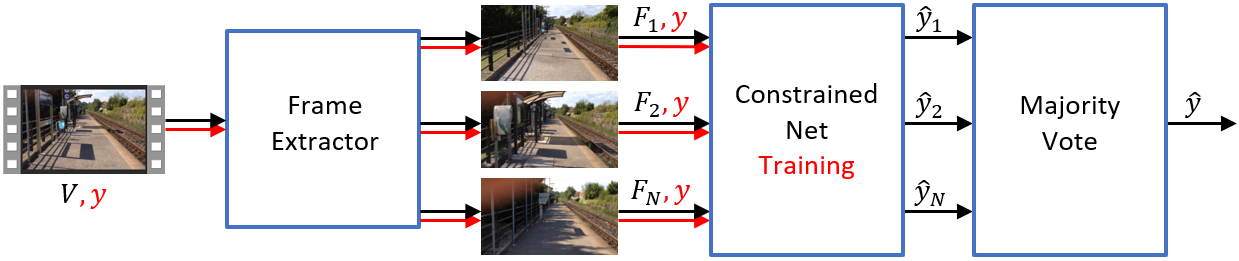, width = 7.5cm}}
    \caption{High level overview of our methodology. During the training process, highlighted in red, $N$ frames are extracted from a video $V$, inheriting the same label $y$ to train the ConstrainedNet. During evaluation, highlighted in black, the ConstrainedNet produces $\hat y_{n}$ labels for $N$ frames to predict the label $\hat y$ of the given video.}
  \label{fig:highlvldiagram}
\end{figure}

\noindent
In Fig.~\ref{fig:highlvldiagram} we illustrate a high level overview of our methodology, which mainly consists of the following steps: i) extraction of frames from the input video; ii) classification of frames by the trained \textit{ConstrainedNet}, and iii) aggregation of frame classifications to produce video-level classifications.

In the following sections, the frame extraction process is explained, as well as the voting procedure. Furthermore, architectural details of the ConstrainedNet are provided. 

\subsection{ConstrainedNet}
\label{sec:constrainednet}
We propose a ConstrainedNet which we make publicly available\footnote{https://github.com/zhemann/vcmi}, 
that we train with deep learning for video device identification based on recent techniques in image \citep{bayar2018constrained} and video \citep{hosler2019video} camera identification. We adapt the constrained ConvNet architecture proposed by \citet{bayar2018constrained} and apply a few modifications. Given the strong variation in video resolutions within the data set that we use, we set the input size of our network equals to the smallest video resolution of $480$ $\times$ $800$ pixels. Furthermore, we increase the size of the first two fully-connected layers from $200$ to $1024$, and most importantly, we extend the original constrained convolutional layer by adapting it for color inputs.

The constrained convolutional layer was originally proposed by \citet{bayar2016deep} and is a modified version of a regular convolutional layer. The idea behind this layer is that relationships exist between neighbouring pixels independent of the scene content. Those relationships are characteristic of a camera device and are estimated by jointly suppressing the high-level scene content and learning connections between a pixel and its neighbours, also referred to as pixel value prediction errors \citep{bayar2016deep}. Suppressing the high-level scene content is necessary to prevent the learning of scene-related features. Therefore, the filters of the constrained convolutional layer are restricted to only learn a set of prediction error filters, and are not allowed to evolve freely. Prediction error filters operate as follows:

\begin{enumerate}
    \item Predict the center pixel value of the filter support by the surrounding pixel values.
    \item Subtract the true center pixel value from the predicted value to generate the prediction error.
\end{enumerate}

\noindent
More formally, \citeauthor{bayar2016deep} placed the following constraints on $K$ filters $\textbf{w}^{(1)}_k$ in the constrained convolutional layer:
\begin{equation}
    \begin{cases}
    \textbf{w}^{(1)}_k(0, 0) = -1\\
    \sum_{m, n \neq 0} \textbf{w}^{(1)}_k(m, n) = 1
    \end{cases}
    \label{eq:constr1d}
\end{equation}

\noindent
where the superscript $^{(1)}$ denotes the first layer of the network, $\textbf{w}^{(1)}_k(m, n)$ is the filter weight at position $(m, n)$ and $\textbf{w}^{(1)}_k(0, 0)$ the filter weight at the center position of the filter support. The constraints are enforced during the training process after the filter's weights are updated in the backpropagation step. The center weight value of each filter kernel is then set to $-1$ and the remaining weights are normalized such that their sum equals $1$.

\subsubsection{Extended Constrained Layer}
The originally proposed constrained convolutional layer only supports gray-scale inputs. We propose an extended version of this layer by allowing it to process inputs with three color channels. Considering a convolutional layer, the main difference between gray-scale and color inputs is the number of kernels within each filter. Whereas gray-scale inputs require one kernel, color inputs require three kernels. Therefore, we modify the constrained convolutional layer by simply enforcing the constraints in Eq.~\ref{eq:constr1d} to all kernels of each filter. The constraints enforced on $K$ $3$-dimensional filters in the constrained convolutional layer can be formulated as follows:

\begin{equation}
    \begin{cases}
    \textbf{w}^{(1)}_{k_j}(0, 0) = -1\\
    \sum_{m, n \neq 0} \textbf{w}^{(1)}_{k_j}(m, n) = 1
    \end{cases}
    \label{eq:constr3d}
\end{equation}

\noindent where $j \in \{1, 2, 3\}$. Moreover, $\textbf{w}^{(1)}_{k_j}$ denotes the $j^{th}$ kernel of the $k^{th}$ filter in the first layer of the ConvNet.

\begin{figure*}[ht]
  \centering
    {\epsfig{file = 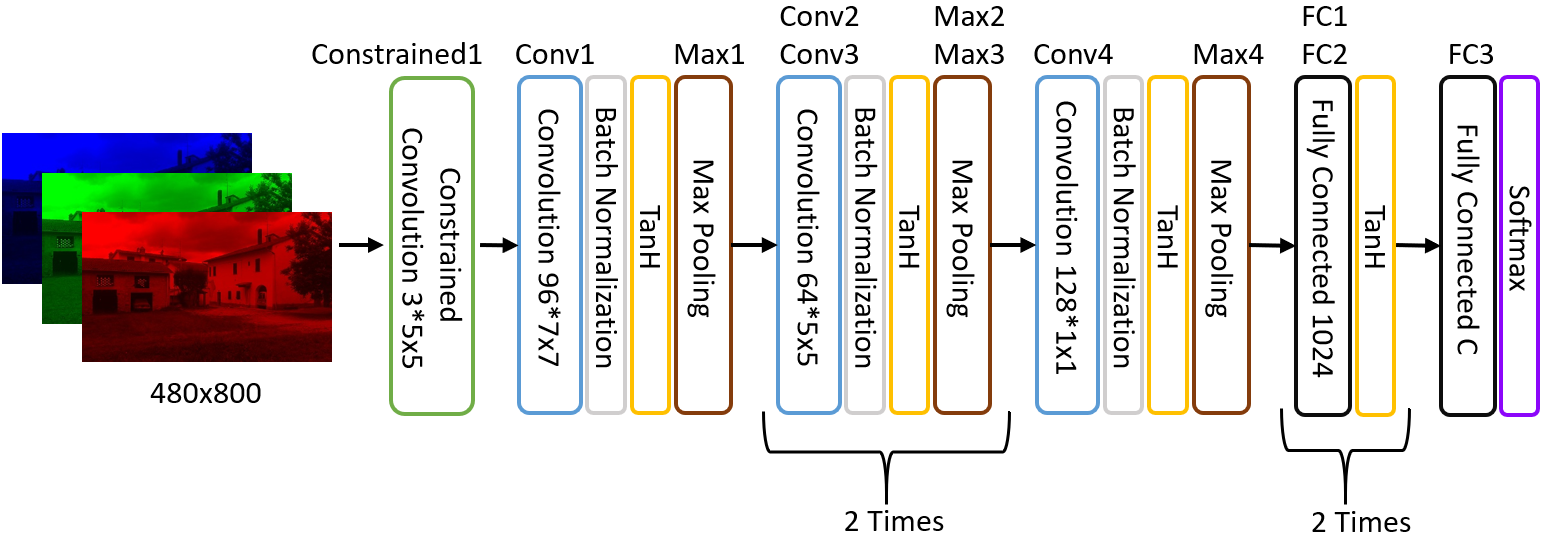, width = 15cm}}
  \caption{Architecture of the proposed ConstrainedNet}
  \label{fig:archconstrainednet}
\end{figure*}

\subsection{Frame Extraction}
\label{sec:frame-extraction}
While other studies extract the first $N$ frames of each video \citep{shullani2017vision}, we extract a given number of frames equally spaced in time across the entire video. For example, to extract $200$ frames from a video consisting of $1000$ frames, we would extract frames $[5$, $10$, .., $1000]$ whereas for a video of $600$ frames we would extract frames $[3$, $6$, .., $600]$. Furthermore, we did not impose requirements on a frame to be selected, in contrast to \cite{hosler2019video}.

\subsection{Voting Procedure}
\label{sec:methods:voting}
The camera device of a video under investigation is identified as follows. We first create the set $I$ consisting of $K$ frames extracted from video $v$, as explained in Section~\ref{sec:frame-extraction}. Then, every input $I_k$ is processed by the (trained) ConstrainedNet, resulting in the probability vector $\textbf{z}_k$. Each value in $\textbf{z}_k$ represents a camera device $c \in C$ where $C$ is the set of camera devices under investigation. We determine the predicted label $\hat y_{k}$ for input $I_k$ by selecting label $y_c$ of the camera device that achieves the highest probability. Eventually, we obtain the predicted camera device label $\hat y_v$ for video $v$ by majority voting on $\hat y_{k}$ where $k \in [1, K]$. 

\section{\uppercase{Experiments and Results}}
\label{sec:exp-results}
\noindent

\subsection{Data set}
\label{sec:dataset}
We used the publicly available VISION data set \citep{shullani2017vision}. It was introduced to provide digital forensic experts a realistic set of digital images and videos captured by modern portable camera devices. The data set includes a total of $35$ camera devices representing $11$ brands. Moreover, the data set consists of $6$ camera models with multiple instances ($13$ camera devices in total), suiting our aim to investigate video camera identification at device level.

The VISION data set consists of $1914$ videos in total which can be subdivided into native versions and their corresponding social media counterparts. The latter are generated by exchanging native videos via social media platforms. There are $648$ native videos, $622$ are shared through YouTube and $644$ via WhatsApp. While both YouTube and WhatsApp apply compression techniques to the input video, YouTube maintains the original resolution while WhatsApp reduces the resolution to a size of $480$ $\times$ $848$ pixels.
Furthermore, the videos represent three different scenarios: \textit{flat}, \textit{indoor}, and \textit{outdoor}. The flat scenario contains videos depicting flat objects such as walls or blue skies, and are often largely similar across multiple camera devices. The indoor scenario comprises videos depicting indoor settings, such as stores and offices, whereas the latter scenario contains videos showing outdoor areas including gardens and streets. Each camera device consists of at least two native videos for every scenario.

\subsubsection{Camera Device Selection Procedure}
Rather than including each camera device of the VISION data set, we selected a subset of camera devices that excludes devices with very few videos. We determined the appropriate camera devices based on the number of available videos and the camera device's model. More explicitly, we included camera devices that met either of the following criteria:

\begin{enumerate}
    \item The camera device contains at least 18 native videos, which are also shared through both YouTube and Whatsapp. \newline
    \item More than one instance of the camera device's brand and model occur in the VISION data set. \newline
\end{enumerate}

\begin{figure}[t]
  \centering
    {\epsfig{file = 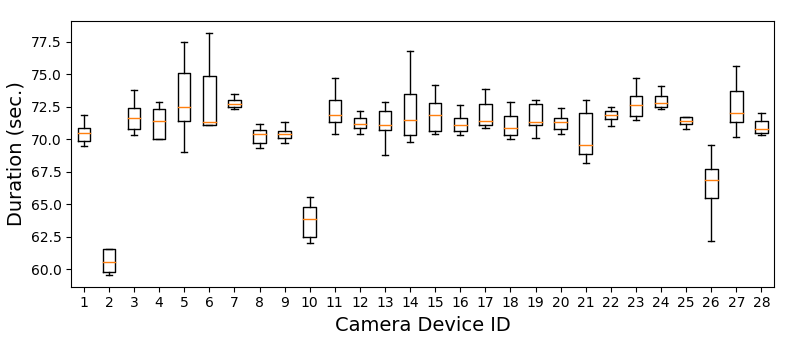, width=7.5cm}}
  \caption{Boxplot showing the means and quartiles of the video durations in seconds for the 28 camera devices. Outliers are not shown in this plot. }
  \label{fig:boxplot-vid-duration}
\end{figure}

\noindent
We applied the first criterion to exclude camera devices that contained very few videos. Exceptions are made for devices of the same brand and model, as indicated by the second criterion. Those camera devices are necessary to exploit video device identification. Furthermore, we excluded the Asus Zenfone 2 Laser camera model as suggested by \cite{shullani2017vision}, resulting in a subset of $28$ camera devices out of $35$, shown in Table~\ref{tbl:devices}. The total number of videos sum up to $1539$ of which $513$ are native and $1026$ are social media versions. In Fig.~\ref{fig:boxplot-vid-duration} an overview of the video duration for the $28$ camera devices is provided.

\begin{table}[t]
\scriptsize
\caption{The set of $28$ camera devices we used to conduct the experiments, of which $13$ are of the same brand and model, indicated by the superscript.}
\label{tbl:devices}
\centering
\begin{tabular}{|l|l|l|l|l|}
\cline{1-2} \cline{4-5}
\textbf{ID} & \textbf{Device} &  & \textbf{ID} & \textbf{Device}   \\ \cline{1-2} \cline{4-5} 
1           & iPhone 4        &  & 15          & Huawei P9         \\ \cline{1-2} \cline{4-5} 
2           & iPhone 4s*       &  & 16          & Huawei P9 Lite    \\ \cline{1-2} \cline{4-5} 
3           & iPhone 4s*       &  & 17          & Lenovo P70A       \\ \cline{1-2} \cline{4-5} 
4           & iPhone 5**       &  & 18          & LG D290           \\ \cline{1-2} \cline{4-5} 
5           & iPhone 5**       &  & 19          & OnePlus 3$^\mathsection$    \\ \cline{1-2} \cline{4-5} 
6           & iPhone 5c$^\dagger$       &  & 20          & OnePlus 3$^\mathsection$    \\ \cline{1-2} \cline{4-5} 
7           & iPhone 5c$^\dagger$       &  & 21          & Galaxy S3 Mini$^\|$   \\ \cline{1-2} \cline{4-5} 
8           & iPhone 5c$^\dagger$       &  & 22          & Galaxy S3 Mini$^\|$   \\ \cline{1-2} \cline{4-5} 
9           & iPhone 6$^{\dagger\dagger}$       &  & 23          & Galaxy S3         \\ \cline{1-2} \cline{4-5} 
10          & iPhone 6$^{\dagger\dagger}$        &  & 24          & Galaxy S4 Mini    \\ \cline{1-2} \cline{4-5} 
11          & iPhone 6 Plus   &  & 25          & Galaxy S5         \\ \cline{1-2} \cline{4-5} 
12          & Huawei Ascend   &  & 26          & Galaxy Tab 3      \\ \cline{1-2} \cline{4-5} 
13          & Huawei Honor 5C &  & 27          & Xperia Z1 Compact \\ \cline{1-2} \cline{4-5} 
14          & Huawei P8       &  & 28          & Redmi Note 3      \\ \cline{1-2} \cline{4-5} 
\end{tabular}
\end{table}

\subsection{Frame-based Device Identification}
\label{sec:exp-results:exp1}
We created balanced training and test sets in the following way. We first determined the lowest number of \textit{native} videos among the $28$ camera devices that we included based on the camera device selection procedure. We used this number to create balanced training and test sets of native videos in a randomized way, maintaining a training-test split of $55\%/45\%$. Moreover, we ensured that per camera device, each scenario (flat, indoor, and outdoor) is represented in both the training and test set. Concerning this experiment, the lowest number of native videos was $13$ through which we randomly picked $7$ and $6$ native videos per camera device for the training and test sets, respectively. Subsequently, we added the WhatsApp and YouTube versions for each native video, tripling the size of the data sets. This approach ensured that a native video and its social media versions always belong to either the training set or test set, but not both. Although the three versions of a video differ in quality and resolution, they still depict the same scene content which could possibly lead to biased results if they were distributed over both the training set and test set. Eventually, this led to $21$ training videos ($7$ native, $7$ WhatsApp, and $7$ YouTube) and $18$ test videos ($6$ native, $6$ WhatsApp, and $6$ YouTube) being included per camera device, resulting in a total number of $588$ training and $504$ test videos. Due to the limited number of available videos per camera device, we did not have the luxury to create a validation set. We extracted $200$ frames from every video following the procedure as described in Section~\ref{sec:frame-extraction}, resulting in a total number of $117,600$ training frames and $100,800$ test frames. Each training frame inherited the same label as the camera device that was used to record the video.

Furthermore, we performed this experiment in two different settings to investigate the performance of our extended constrained convolutional layer. In the first setting we used the ConstrainedNet as explained in Section~\ref{sec:constrainednet}, whereas we removed the constrained convolutional layer in the second setting. We refer to the network in the second setting as the \textit{UnconstrainedNet}.

We trained both the ConstrainedNet and the UnconstrainedNet for $30$ epochs in batches of $128$ frames. We used the stochastic gradient descent (SGD) during the backpropagation step to minimize the categorical cross-entropy loss function. To speed up the training time and convergence of the model, we used the momentum and decay strategy. The momentum was set to $0.95$ and we used a learning rate of $0.001$ with step rate decay of $0.0005$ after every training batch. The training took roughly $10$ hours to complete for each of both architectures\footnote{We used the deep learning library Keras on top of Tensorflow to create, train, and evaluate the  ConstrainedNet and UnconstrainedNet. Training was performed using a Nvidia Tesla V100 GPU.}.

We measured the performance of the ConstrainedNet and the UnconstrainedNet by calculating the video classification accuracy on the test set. After every training epoch we saved the network's state and calculated the video classification accuracy as follows: 
\begin{enumerate}
    \item Classify each frame in the test set.
    \item Aggregate frame classifications per video.
    \item Classify the video according to the majority vote as described in Section~\ref{sec:methods:voting}.
    \item Divide the number of correctly classified test videos by the total number of test videos.
\end{enumerate}

In addition to the test accuracy, we also calculated the video classification accuracy for the different scenarios (flat, indoor, outdoor) and versions (native, WhatsApp, YouTube). The different scenarios are used to exploit the extraction of noise pattern features and to determine the influence of high-level scene content. The different video versions are used to investigate the impact of different compression techniques.

\subsubsection{Results}
\begin{figure}[ht]
  \centering
    {\epsfig{file = 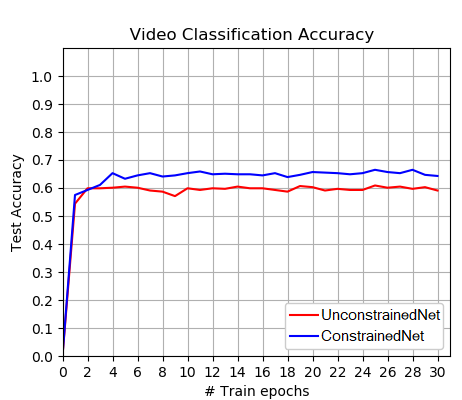, width = 7cm}}
  \caption{Progress of the test video classification accuracy for the ConstrainedNet and UnconstrainedNet.}
  \label{fig:avg-constrained-vs-conv}
\end{figure}

\begin{figure*}[ht]
  \centering
    {\epsfig{file = 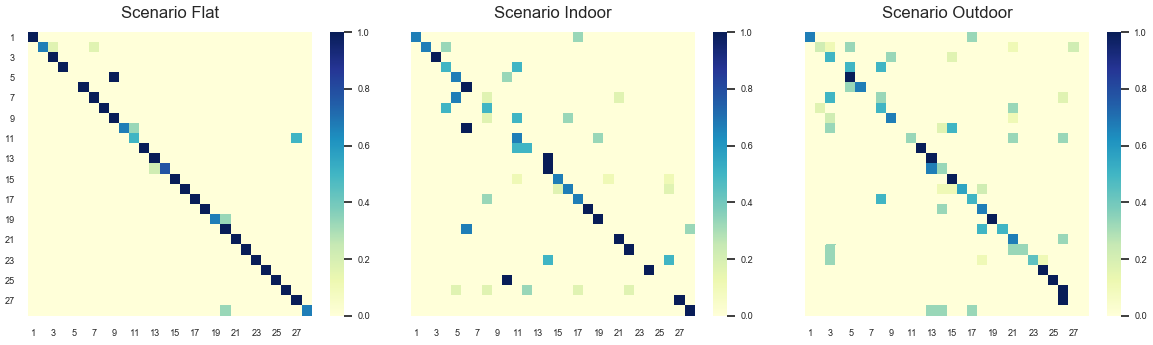, width = 16cm}}
  \caption{Confusion matrices for $28$ camera devices showing the normalized classification accuracies achieved for the scenarios flat, indoor, and outdoor.}
  \label{fig:exp1:cm-comb-scenario-small}
\end{figure*}

In Fig.~\ref{fig:avg-constrained-vs-conv} we show the progress of the test accuracy for the ConstrainedNet and the UnconstrainedNet. It can be observed that the performance significantly improved by the introduction of the constrained convolutional layer. The ConstrainedNet achieved its peak-performance after epoch 25 with an overall accuracy of $66.5$\%. Considering the accuracy per scenario, we observed $89.1\%$ accuracy for flat scenario videos, $53.7\%$ for indoor scenarios, and $55.2\%$ accuracy for the outdoor. Results for the individual scenarios are shown in Fig.~\ref{fig:exp1:cm-comb-scenario-small}.

\begin{figure*}[ht]
  \centering
    {\epsfig{file = 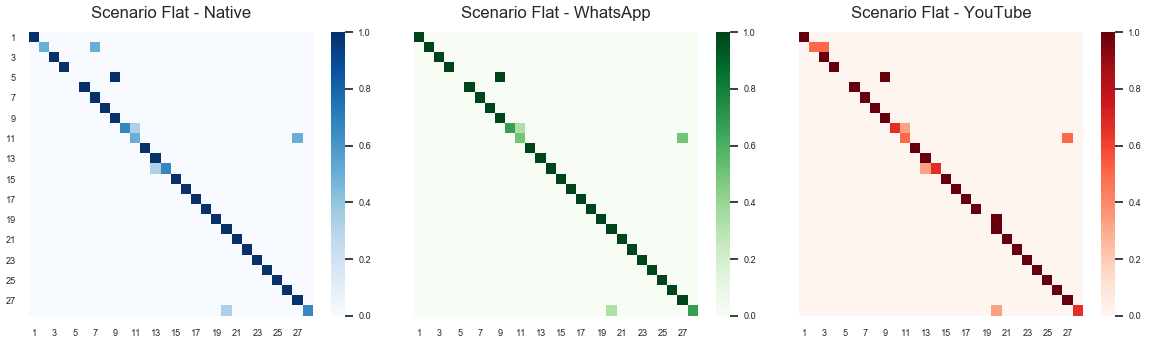, width = 16cm}}
  \caption{Confusion matrices for $28$ camera devices showing the normalized classification accuracies achieved on flat scenario videos from the perspective of the native videos and their social media versions.}
  \label{fig:exp1:cm-comb-flat-version-small}
\end{figure*}

Since the flat scenario videos achieved a significantly higher classification accuracy compared to the others, we limited ourselves to this scenario during the investigation of how different compression techniques would affect the results. In Fig.~\ref{fig:exp1:cm-comb-flat-version-small} we show the confusion matrices that the ConstrainedNet achieved from the perspective of different compression techniques (i.e. native, WhatsApp and YouTube). We achieved $89.7\%$ accuracy on the native versions, $93.1\%$ on the WhatsApp versions, and $84.5\%$ on the YouTube ones.

\section{\uppercase{Discussion}}
\label{sec:discussion}
\noindent
From the results in Fig.~\ref{fig:avg-constrained-vs-conv} it can be observed that the constrained convolutional layer significantly contributes to the performance of our network. This suggests this layer is worth further investigating its potential in light of digital forensic problems on color inputs. Furthermore, in Fig.~\ref{fig:exp1:cm-comb-scenario-small} it is shown  that the classification accuracy greatly differs between the three scenarios; flat, indoor, and outdoor.  Whereas the indoor and outdoor scenarios achieve accuracies of $53.7\%$ and $55.2\%$, respectively, we observe an accuracy of $89.1\%$ for the flat scenario videos. Given the high degree of similarity between flat scenario videos from multiple camera devices, the results suggest that the ConstrainedNet has actually extracted characteristic noise pattern features for the identification of source camera devices. The results also indicate that indoor and outdoor scenario videos are less suitable to extract noise pattern features for device identification. This difference could lie in the absence or lack of video homogeneity. Compared to the flat scenario videos, indoor and outdoor scenario videos typically depict a constantly changing scene. As a consequence, the dominant features of a video frame are primarily scene-dependent, making it significantly harder for the ConstrainedNet to extract the scene-independent features, that is, the noise pattern features. 

Fig.~\ref{fig:exp1:cm-comb-flat-version-small} shows that our methodology is robust against the compression techniques applied by WhatsApp and YouTube. While we observe an accuracy of $89.7\%$ for the native versions of flat scenario videos, we observe accuracies of $93.1\%$ and $84.5\%$ for the WhatsApp and YouTube versions, respectively. The high performance of WhatsApp versions could be due to the similarity in size (i.e. resolution) between WhatsApp videos and our network's input layer. As explained in Section~\ref{sec:dataset}, the WhatsApp compression techniques resize the resolution of a video to the size of $480 \times 848$ pixels, becoming nearly identical to the network's input size of $480 \times 800$ pixels. This is in contrast to the techniques applied by YouTube, which respect the original resolution.

These results indicate that the content homogeneity of a video frame plays an important role in the classification process of videos. Therefore, we suggest to search for homogeneous patches within each video frame, and only use those patches for the classification of a video. This would limit the influence of scene-related features, forcing the network to learn camera device specific features.

By proposing this methodology we aim to support digital forensic experts in order to identify the source camera device of digital videos. We have shown that our approach is able to identify the camera device of a video with an accuracy of $89.1\%$. This accuracy further improves to $93.1\%$ when considering the WhatsApp versions. Although the experiments were performed on known devices, we believe this work could be extended by matching pairs of videos of known and unknown devices too. In that case, the ConstrainedNet may be adopted in a two-part deep learning solution where it would function as the feature extractor, followed by a similarity network (e.g. Siamese Networks) to determine whether two videos are acquired by the same camera device. To the best of our knowledge, this is the first work that addresses the task of device-based video identification by applying deep learning techniques. 

In order to further improve the performance of our methodology, we believe it is worth investigating the potential of patch-based approaches wherein the focus lies on the homogeneity of a video. More specifically, only homogeneous patches would be extracted from frames and used for the classification of a video. This should allow the network to better learn camera device specific features, leading to an improved device identification rate. Moreover, by using patches the characteristic noise patterns remain unaltered since they do not undergo any resize operation. In addition, this could significantly reduce the complexity of the network, requiring less computational effort. Another aspect to investigate would be the type of voting procedure. Currently, each frame always votes for a single camera device even when the ConstrainedNet is highly uncertain about which device the frame is acquired by. To counteract this problem, voting procedures could be tested that take this uncertainty into account. For example, we could require a certain probability threshold to vote for a particular device. Another example would be to select the camera device based on the highest probability after averaging the output probability vectors of frames aggregated per video. 

\section{\uppercase{Conclusion}}
\label{sec:conclusion}
\noindent
Based on the results that we achieved so far, we draw the following conclusions. The extended constrained convolutional layer contributes to increase in performance. Considering the different types of videos, the proposed method is more effective for videos with flat (i.e. homogeneous) content, achieving an accuracy of $89.1\%$. In addition, the method shows to be robust against the WhatsApp and YouTube compression techniques with accuracy rates up to $93.1\%$ and $84.5\%$, respectively. 

\section*{\uppercase{Acknowledgements}}
\noindent 
We thank the Center for Information Technology of the University of Groningen for their support and for providing access to the Peregrine high performance computing cluster. This research has been funded with support from the European Commission under the 4NSEEK project with Grant Agreement 821966. This publication reflects the views only of the authors, and the European Commission cannot be held responsible for any use which may be made of the information contained therein.

\vfill

\bibliographystyle{apalike}
{\small
\bibliography{example}}

\end{document}